\documentclass[10pt, conference]{IEEEtran}
\usepackage[left=0.625in,right=0.625in,top=0.75in,bottom=0.99in]{geometry}
\IEEEoverridecommandlockouts
\usepackage{cite}
\usepackage{amsmath,amssymb,amsfonts}
\usepackage{graphicx}
\usepackage{color}
\usepackage{subfig}
\usepackage{stackengine}
\usepackage{comment}
\usepackage[normalem]{ulem}
\usepackage{caption}
\usepackage{tabularx}
\usepackage{bm}
\usepackage{dsfont}
\usepackage{mathtools}
\usepackage{lettrine}

\newcommand{\noteBK}[1]{\textcolor{purple}{[{\bf #1}]}}
\usepackage[hidelinks]{hyperref} 

\usepackage[flushleft]{threeparttable} 
\usepackage{booktabs}

\usepackage{textcomp}
\usepackage{xcolor}

\usepackage[linesnumbered, ruled]{algorithm2e}
\SetKwRepeat{Do}{do}{while}%

\SetCommentSty{mycommfont}
\usepackage{algpseudocode}

\def\BibTeX{{\rm B\kern-.05em{\sc i\kern-.025em b}\kern-.08em
    T\kern-.1667em\lower.7ex\hbox{E}\kern-.125emX}}
    
\usepackage{orcidlink}

\usepackage{xspace}
\usepackage{ifthen}
\usepackage[inline]{enumitem}

\usepackage{algcompatible}

\newboolean{showcomments}
\setboolean{showcomments}{true}
\ifthenelse{\boolean{showcomments}}
{
}

\usepackage{subfig}
\usepackage{fancyhdr}

\begin{document}

\title{Spatiotemporal Semantic V2X Framework for Cooperative Collision Prediction}

\author{Murat Arda Onsu$^1$, Poonam Lohan$^1$, Burak Kantarci$^1$, Aisha Syed$^2$, Matthew Andrews$^2$, Sean Kennedy$^2$
\thanks{$^{1}$School of Electrical Engineering and Computer Science,
University of Ottawa, Ottawa, ON, Canada. 
Emails: \{monsu022, ppoonam, burak.kantarci\}@uottawa.ca}
\thanks{$^2$Nokia Bell Labs, 600 March Road,
Kanata, ON K2K 2E6, Canada. \newline 
Emails: \{aisha.syed, matthew.andrews, sean.kennedy\}@nokia-bell-labs.com}
}

\maketitle
\begin{abstract}
Intelligent Transportation Systems (ITS) demand real-time collision prediction to ensure road safety and reduce accident severity. Conventional approaches rely on transmitting raw video or high-dimensional sensory data from roadside units (RSUs) to vehicles, which is impractical under vehicular communication bandwidth and latency constraints. In this work, we propose a semantic V2X framework in which RSU-mounted cameras generate spatiotemporal semantic embeddings of future frames using the Video Joint Embedding Predictive Architecture (V-JEPA). Afterwards, These embedding future frames are utilized to estimate the likelihood collision for alert the prevention system at vehicle side.  
To evaluate the system, we construct a digital twin of an urban traffic environment enabling the generation of diverse traffic scenarios with both safe and collision events. The embeddings of the future frame, extracted from V-JEPA, capture task-relevant traffic dynamics and are transmitted via V2X links to vehicles, where a lightweight classifier decode them to predict imminent collisions. By transmitting only semantic embeddings instead of raw frames, the proposed system significantly reduces communication payload while maintaining predictive accuracy. Experimental results demonstrate that the proposed framework with an appropriate processing method achieves 92\% accuracy and 8\% F1-score improvement for collision prediction. It also 
reduces transmission payload by five orders of magnitude compared to raw video fulfilling the latency constraints for safe driving. This validates the potential of semantic V2X communication to enable cooperative, real-time collision prediction in ITS.
\end{abstract}


\section{Introduction} \label{sec:1}

Achieving proactive collision prediction is essential for enhancing road safety through early warnings and coordinated preventive actions in Intelligent Transportation Systems (ITS) \cite{1_ITS}. However, conventional approaches that transmit raw video streams \cite{1_S} or high-dimensional sensor data from roadside units (RSUs) to vehicles are impractical under vehicular communication (V2X) constraints due to limited bandwidth and stringent latency requirements. These limitations highlight the need for communication-efficient, task-oriented frameworks that convey only semantically relevant information to support predictive safety functions in connected and autonomous vehicles.

Semantic communication (SemComm) addresses this by transmitting task-relevant features instead of raw sensory data \cite{SEMCOM}. In vehicular contexts, this enables roadside units (RSUs) to extract compressed semantic representations that capture the dynamics most relevant to safety, and vehicles to interpret them for decision-making. Crucially, if these semantic representations are predictive rather than descriptive, vehicles can receive early warnings of imminent collisions rather than delayed notifications of ongoing accidents.
To evaluate such systems, digital twins play a key role in generating realistic, controllable, and reproducible traffic scenarios. In this work, we employ the Quanser Interactive Labs (QLabs) platform as a digital twin of an urban traffic system. This environment provides diverse scenarios including intersections and roundabouts, where both safe and collision cases can be simulated. The digital twin thus enables systematic video-data generation for training and evaluating predictive SemComm frameworks.

Building on this foundation, we propose a novel semantic V2X collision prediction framework which departs from conventional perception-centric V2X approaches by performing future-oriented semantic prediction.
RSUs equipped with cameras employ the Video Joint Embedding Predictive Architecture (V-JEPA) as the semantic encoder \cite{2}. V-JEPA does not merely summarize observed frames; it predicts the embeddings of \textit{future frames}, capturing the evolving motion patterns and vehicle interactions that lead to collisions. These predicted embeddings are transmitted to vehicles via V2X links, where a classifier acts as a semantic decoder, refining the embeddings to predict whether a collision is imminent. By focusing on predictive embeddings, the system moves beyond detection toward true anticipation of dangerous events. The main contributions of this work are summarized as follows:
\begin{itemize}
    \item A high-quality video dataset is generated using a QLabs digital twin, featuring diverse urban scenarios labeled as collision or safe, supporting systematic evaluation of predictive collision avoidance.
    
    \item We evaluate post-processing methods such as heatmaps, binary masks, and hybrids to enhance task-relevant features and assess their effect on V-JEPA’s representation quality and predictive accuracy.
    
    \item V-JEPA is employed to encode future-frame embeddings at RSUs, which are transmitted via V2X and decoded by  a lightweight classifier in vehicles to identify high-risk regions for collision prediction.
\end{itemize}

This work shows that combining digital twins, SemComm, and spatiotemporal embeddings can enable predictive collision avoidance in vehicular networks, achieving 92\% accuracy and  8\% F1-score improvement in collision prediction with five orders of magnitude reduction in transmission (Tx) payload, supporting real-time latency constraint V2X communication. 

\section{Related Works} \label{sec:2}
Collision prediction and avoidance have been explored from various perspectives. In \cite{8}, a deep learning model based on convolutional neural networks (CNNs) analyzes surveillance video for traffic congestion and accident detection, addressing the limits of manual observation and conventional learning methods. Similarly, the work in \cite{9} uses ensemble models (Random Forest, XGBoost, L-GBM, CatBoost) with SHAP-based interpretability to assess accident severity and contributing factors. However, these studies mainly emphasize post-event analysis rather than proactive prediction of collisions.

Big data frameworks have also been leveraged in \cite{10}, where large-scale UK traffic datasets were processed using Apache Spark with Decision Trees, Random Forest, and Naïve Bayes to predict accident severity under diverse environmental conditions. Meanwhile, \cite{11} introduced the I3D-CONVLSTM2D model that integrates spatial and temporal features from RGB and optical flow data for real-time accident detection on edge devices, emphasizing computational efficiency. Although spatiotemporal features improve recognition, these methods detect accidents as they occur, without forecasting future risk.

Recently, the use of LLMs has been explored for traffic video analysis. In \cite{12}, the VAD-LLaMA framework combined video anomaly detection with natural language explanations, using a Long-Term Context (LTC) module and instruction-tuned Video-LLaMA to localize and explain anomalies. Similarly, \cite{13} proposed a hybrid framework combining Bayesian Generalized Extreme Value (GEV) models with ARIMA/ARIMAX forecasting and AI-based video analytics to estimate short-term crash risks at intersections. While these approaches improve interpretability, they rely heavily on temporal correlations in data streams rather than predictive spatiotemporal embeddings tailored for proactive vehicular safety.

YOLO-based detection frameworks have gained attention for traffic analysis. In \cite{14}, YOLOv8 and YOLO11x models are trained on 15,000 traffic images for real-time accident and speeding-vehicle detection. For proactive collision avoidance, the work in \cite{15} integrates CARLA simulations with YOLOv7-based object detection, while the study in \cite{16} enhances small-object detection and near-miss prediction through instance segmentation, IPM, and tracking modules. Although effective for object-level detection, these methods still rely on transmitting large volumes of visual data for centralized processing.

In contrast, this work introduces a V2X semantic communication framework where RSUs use V-JEPA to generate predictive spatiotemporal embeddings of future frames, enabling anticipation of collision-prone motion patterns. Rather than transmitting raw video, semantic embeddings are shared with vehicles for decoding via a lightweight classifier, achieving real-time and bandwidth-efficient collision prediction.

\section{Methodology}
This work develops a semantic V2X framework in which RSUs equipped with surveillance cameras and V-JEPA act as semantic encoders, and vehicles act as semantic decoders for predictive collision avoidance. Unlike image-based approaches, which provide only static snapshots, video data captures both spatial and temporal dynamics, enabling richer modeling of vehicle interactions, motion patterns, and speed estimation. V-JEPA is employed at the RSU to shift predictive modeling from pixel-level forecasting to the representation space. Instead of reconstructing raw frames, V-JEPA generates compact spatiotemporal embeddings of \textit{future frames}, thereby anticipating anomalous motion patterns that precede collisions. These predicted embeddings are transmitted as semantic messages over V2X links to vehicles, where a classifier decode them to determine whether a collision is imminent or the driving condition is safe. By transmitting embeddings rather than raw video, the system reduces both communication payload and computational workload while preserving predictive accuracy.

\begin{figure*}
    \centering
    \includegraphics[width=0.7\linewidth]{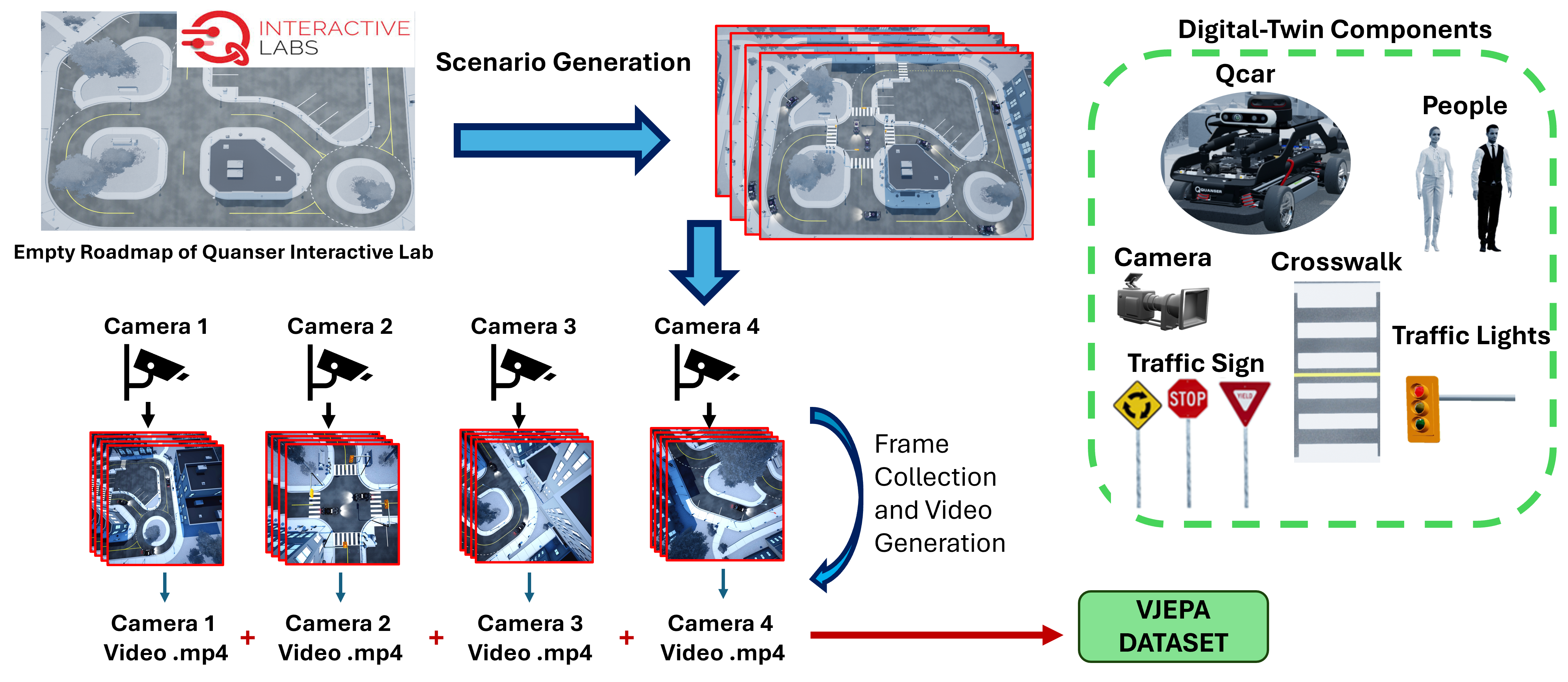}
    \caption{Data Collection Using Digital-Twin (Quanser Interactive Lab)}
    \label{fig:1}
\end{figure*}

\subsection{Dataset Generation}

Dataset generation for the V-JEPA modules consists of two steps: (i) video collection using a digital twin environment, and (ii) post-processing to emphasize task-relevant regions.  

\subsubsection{Video Generation using QLabs} \label{qlab}
To obtain a realistic and diverse dataset, the QLabs platform is employed as a digital twin of an urban traffic environment \cite{1_Q}. QLabs models intersections, roundabouts, two-way roads, pedestrians, traffic lights, and autonomous vehicles, enabling flexible and reproducible scenario creation. The Qlabs environment is programmed to generate both safe driving and collision events, observed from four RSU-mounted surveillance cameras covering different regions of the environment. Simulation provides 4 different type of regions: 4-way junction, 3-way junction, side roads and roundabout. There are total 500 video clips and each frame is sampled every 50 ms, producing 20 FPS (Frames per second) of varying lengths depending on scenario duration. This allows for reproducible traffic scene generation and controlled experimentation for predictive safety analysis. \figurename \ref{fig:1} depicts the process of video-dataset generation.

\subsubsection{Video Post-Processing} \label{post-processing}

\begin{figure}
    \centering
    \includegraphics[width=0.50\linewidth]{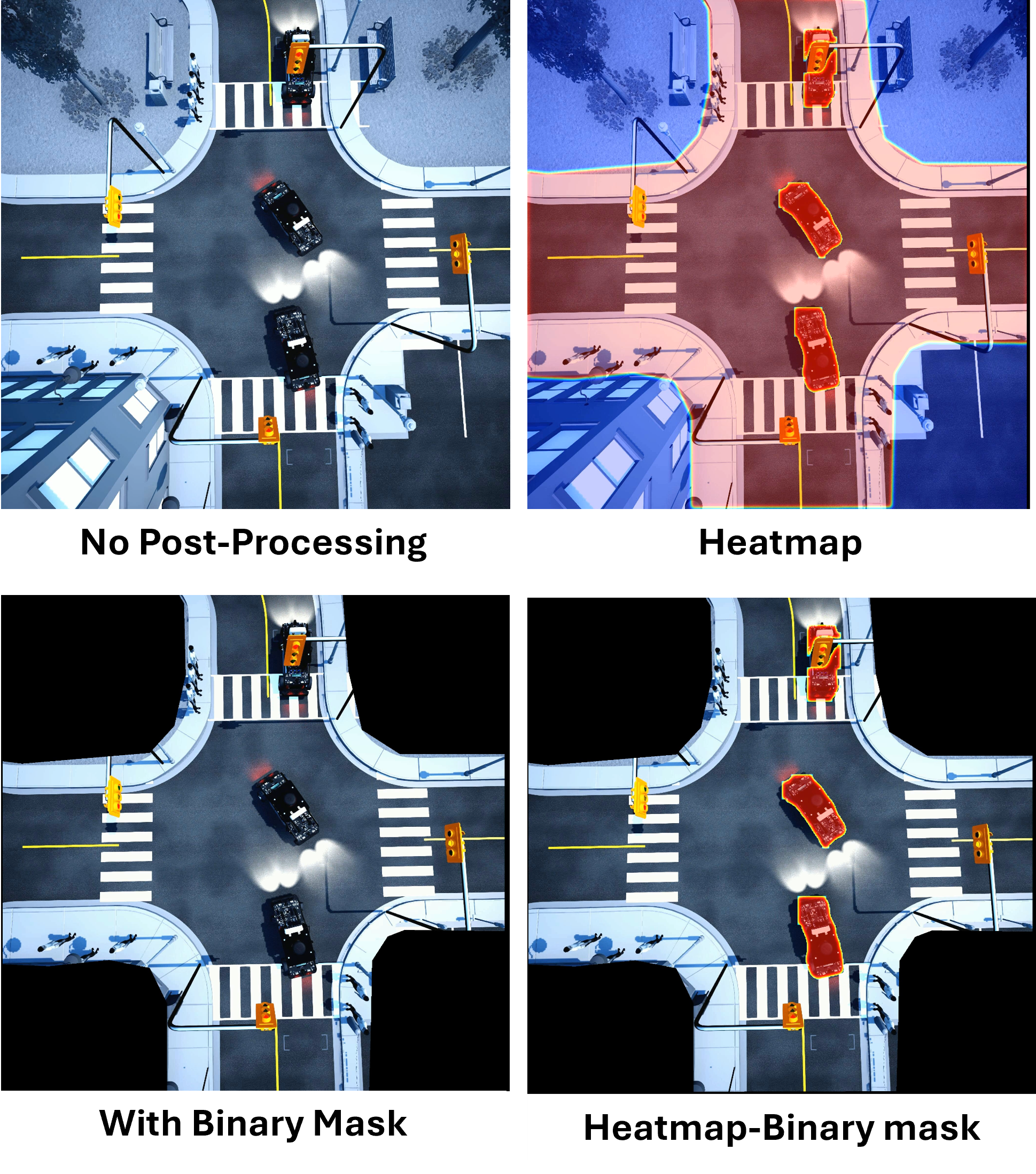}
    \caption{Different Post-processing Methods for Video Dataset}
    \label{fig:2}
\end{figure}

The collected video clips are subjected to a post-processing stage designed to emphasize task-relevant regions and simplify downstream learning. YOLOv11 \cite{1_Y} is employed to detect vehicles, after which three techniques are applied: (i) heatmaps, (ii) binary road masks, and (iii) a hybrid of heatmaps and binary masking. In the heatmap approach, vehicle coordinates are highlighted first, followed by road regions, while non-road areas such as surrounding trees and infrastructure are down-weighted. The binary masking approach assigns zero weight to all pixels outside the road, whereas the hybrid method further highlights vehicle coordinates while suppressing non-road regions entirely. These techniques highlight dynamic objects  and road areas while reducing background redundancy with examples shown in \figurename \ref{fig:2}.

\subsection{Semantic Encoder: V-JEPA at RSU}
The V-JEPA model \cite{2} is deployed at RSUs to generate predictive semantic embeddings. Unlike conventional predictive models, V-JEPA forecasts embeddings of masked \textit{future frame} rather than reconstructing pixels, thereby focusing on task-relevant semantics for downstream prediction.  

\subsubsection{V-JEPA Pretraining with Masked Modeling} \label{pretrain}
\begin{figure*}
    \centering
\includegraphics[width=0.85\linewidth]{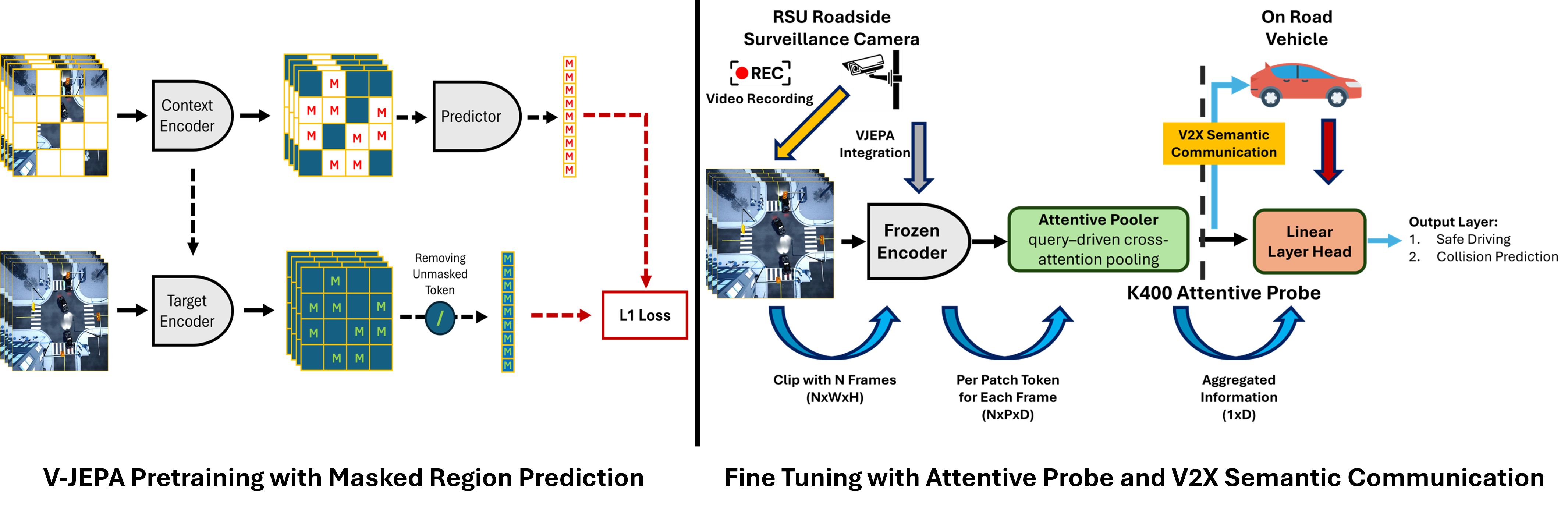}
    \caption{V-JEPA Pretraining and Fine-Tuning for Collision Prediction}
    \label{fig:8}
\end{figure*}

As shown in \figurename~\ref{fig:8}, V-JEPA is pretrained in a self-supervised fashion using masked prediction. A video clip is divided into non-overlapping spatiotemporal patches, where each frame of spatial size $H \times W$ is partitioned into $P=\tfrac{H}{p} \times \tfrac{W}{p}$ patches based on a patch size of $p \times p$. The target encoder processes the full unmasked clip, while the context encoder encodes a masked version. Each patch is projected into a $D$-dimensional embedding vector, resulting in a per-frame embedding tensor of size $P \times D$. A predictor network then generates embeddings for the masked tokens, which are trained against the corresponding target encoder embeddings using an $L_1$ loss. This enables the model to learn predictive spatiotemporal representations directly in the embedding space rather than at the pixel level.

\subsubsection{Fine-Tuning with Attentive Probe} \label{finetune}
For downstream collision prediction, the pretrained V-JEPA encoder takes as input a video clip of dimension $N \times W \times H$, where $N$ is the number of frames in the clip. The clip is flattened into a sequence of non-overlapping spatiotemporal patches, producing token embeddings of shape $N \times P \times D$, where $P$ is the number of patches per frame and $D$ is the embedding dimension. These frozen encoder parameters preserve the rich self-supervised features learned during pretraining. On top of the encoder outputs, The attentive probe employs a query-driven cross-attention mechanism, where the query vector acts as a learnable global token that attends to the most informative spatial and temporal regions in the video. Formally, let the encoder output token embeddings be $\mathbf{Z} \in \mathbb{R}^{(N \cdot P) \times D}$. The attentive probe introduces a learnable query matrix $\mathbf{Q} \in \mathbb{R}^{Q \times D}$, with $Q=1$ in this setup, to perform cross-attention as follows:

\begin{equation}\small
\mathbf{z}_{\text{attn}}=
\text{Attn}(\mathbf{Q}, \mathbf{Z}) = \text{softmax}\left(\frac{\mathbf{Q}\mathbf{Z}^\top}{\sqrt{D}}\right)\mathbf{Z},
\end{equation}

where $\mathbf{Q}$ serves as the query, and $\mathbf{Z}$ provides both keys and values.
The resulting weighted aggregation yields a single compact embedding $\mathbf{z}_{\text{attn}} \in \mathbb{R}^{1 \times D}$, which captures the most salient spatiotemporal dynamics relevant to collision prediction.
 This pooled embedding is then passed to a lightweight classifier to infer one of two states: \textit{collision} or \textit{safe driving}. By transforming raw video data ($N \times W \times H$) into a compact embedding $1 \times D$, the framework enables efficient semantic Tx while maintaining predictive accuracy for V2X communication.

\subsection{Semantic Decoder at Vehicle}

At the vehicle side, the received embeddings are directly processed by the semantic decoder, which consists of a lightweight classifier. Vehicles receive embeddings from RSUs, that are refined through query-driven cross-attention, concentrating on the most critical spatial and temporal cues that may indicate anomalous motion or collision risk. The resulting compact representation is then fed into a linear classifier to produce a binary decision: \textit{collision} or \textit{safe driving}.  

Since the decoder operates only on low-dimensional embeddings rather than raw video streams, the computational complexity of the decoder is modest (explained in Sec. \ref{Complex})
This makes the design suitable for real-time inference on resource-constrained vehicular hardware, while also significantly reducing communication overhead in the V2X link. The division of computation heavy embedding extraction at the RSU and lightweight decoding at the vehicle ensures scalability and practicality for latency-sensitive V2X networks.

\subsection{Transmission Efficiency in V2X Semantic Communication} \label{section:embedding}

A major advantage of the proposed framework lies in its ability to drastically reduce Tx payloads compared to conventional raw video communication. Let each video clip contain $N$ frames of original spatial size $H_o \times W_o$ with 3 RGB channels, where $(H_o, W_o)$ denote the preprocessed frame dimensions. The total raw video payload can be expressed as:
\begin{equation}\small
S_{\text{raw}} = N \cdot H_o \cdot W_o \cdot 3 \quad \text{bytes}.
\end{equation}


In contrast, the semantic embeddings generated by the V-JEPA encoder at the RSU are significantly more compact. Each clip produces embeddings of size $N \times P \times D$, where $P$ is the number of spatiotemporal patches per frame and $D$ is the embedding dimension. Then, an attentive probe aggregates these tokens into a single compact representation of size $1 \times D$ using a query-driven cross-attention mechanism. This process selectively emphasizes the most informative spatial and temporal regions, resulting in a lightweight semantic vector suitable for V2X transmission. The payload corresponding to the transmitted semantic embedding is therefore:
\begin{equation}\small
S_{\text{sem}} = 1\times D \times b \quad \text{bytes},
\end{equation}
where $b$ denotes the number of bytes per element ($b=4$ for FP32, $b=2$ for FP16, $b=1$ for INT8).
The semantic compression ratio can be defined as $\mathcal{R} = {S_{\text{raw}}}/{S_{\text{sem}}}.$


\subsection{Complexity Analysis of Frozen Encoder with Attentive Probe}\label{Complex}

Consider a video clip composed of $N$ frames, each of spatial size $H \times W$.  
The spatiotemporal tokenizer divides the clip into non-overlapping patches of size $(t_p \times p \times p)$, where $t_p$ is the temporal stride and $p$ is the spatial stride. This process yields a token sequence of length, $L = \frac{N}{t_p} \cdot \frac{H}{p} \cdot \frac{W}{p}$,
where $L$ denotes the total number of spatiotemporal tokens fed into the encoder.
\subsubsection{Frozen Encoder Cost}
Let the encoder embedding dimension be $D$, the number of Transformer blocks be $L_e$, and the multi-layer perceptron (MLP) feed-forward expansion ratio be $r$. Each Transformer block consists of a multi-head self-attention (MSA) layer followed by an MLP. The floating-point operations (FLOPs) per block can be approximated as $F_{\text{block}} = 2L^2D + (4 + 2r)LD^2$,
where the first term corresponds to attention computations ($\mathcal{O}(L^2D)$) and the second term represents the projection and feed-forward operations ($\mathcal{O}(LD^2)$).  
The total cost of frozen encoder over all $L_e$ layers is therefore:
\begin{equation}\small
F_{\text{enc}} = L_e \bigl( 2L^2D + (4 + 2r)LD^2 \bigr).
\end{equation}
During inference, the peak activation memory of the encoder (without gradient storage) scales as:
$\mathcal{M}_{\text{enc}} = \mathcal{O}(LD)$,
dominated by token embeddings and attention buffers.

\subsubsection{Attentive Probe and Classifier Cost}

The attentive probe aggregates the encoder’s token sequence using a single-query cross-attention mechanism, followed by a small MLP and a linear classification layer.  
The computational cost of the probe can be expressed as:
\begin{equation}\small
F_{\text{probe}} = 3LD^2 + 2LD + 3D^2 + DC,
\end{equation}
where $C$ denotes the number of output classes (e.g., $C=2$ for collision and safe driving).  
However, in practice, the encoder output is frozen and the probe operates with a \textit{single query vector} rather than performing full self-attention over all $L$ tokens to capture a unified global representation.  
Consequently, the cost associated with key and value projections ($\mathcal{O}(LD^2)$) is computed once during encoding and not repeated during probing.  
The probe therefore performs only lightweight attention ($\mathcal{O}(LD)$) and projection operations ($\mathcal{O}(D^2)$), resulting in an overall complexity that scales as:
\begin{equation}
\mathcal{O}(D^2 + DC),\small
\end{equation}
where the $D^2$ term arises from the linear transformations within the probe and the $DC$ term from the final classification layer.  
This single-query design substantially reduces the number of token interactions, making the decoder computationally efficient and suitable for real-time execution on vehicular hardware.

\subsubsection{End-to-End Computational Cost}

For a single video clip and spatiotemporal view,  total forward-pass cost is given by:
\begin{equation}\small
F_{\text{total}} = F_{\text{enc}} + F_{\text{probe}}.
\end{equation}
If $V$ different spatiotemporal views (e.g., crops or augmentations) are processed per clip, the total cost scales linearly with $V$.  
Given a device with computational throughput $\phi$ (in FLOPs/s) and per-view input/output latency $t_{\text{I/O}}$, the total inference time can be approximated as $
t_{\text{infer}} = {F_{\text{total}}}/{\phi} + V \, t_{\text{I/O}}.$
Hence, the encoder dominates the total computational cost, while the attentive probe and classifier introduce only minor overhead ($\mathcal{O}(D^2 + DC)$).  
This computational asymmetry justifies deploying the encoder at the RSU (with higher compute resources) and the lightweight decoder on the vehicle side.

\section{Performance Evaluation}


\subsection{Evaluation Setup and Parameters}

As mentioned in Section \ref{qlab}, four different cameras are used in the QLabs platform to collect consecutive frames at 50 ms intervals, which are then used to generate video clips at 20 FPS. In total, 500 traffic scenario clips have been collected, with 115 including collisions and 385 consisting of safe driving\footnote{Urban Mobility Research Video Dataset, Available at: \url{https://dx.doi.org/10.21227/jazm-2z18}. Generated with Quanser Interactive Lab: \url{(https://www.quanser.com/products/quanser-interactive-labs/)}}.  The clips' durations vary depending on the simulation runtime. The simulation automatically finalizes itself once a collision occurs. The long clips' lengths are adjusted to a maximum of 64 frames by removing from the beginning to make it adaptable for V-JEPA training parameters. For various post-processing methods, explained in Section \ref{post-processing}, a previously trained YOLOv11 model is employed from \cite{6}.

For the V-JEPA model, a ViT-Huge (ViT-H) encoder pretrained with VideoMix2M via masked modeling self-supervised learning \cite{2} is employed, as described in Section~\ref{pretrain}. Each video clip is resized to 384×384 before encoding, with a patch size of 16×16 and an embedding dimension of 1280 used in the SemComm framework. For task-specific fine-tuning (Section~\ref{finetune}), the K400 attentive probe is trained for 40 epochs with a batch size of 8, using the AdamW optimizer and an initial learning rate of 0.001. The output layer includes two nodes, namely the Safe Driving and Collision Probability, and the training parameters are summarized in Table~\ref{tab:parameters}.

\begin{table}[]
\renewcommand{\arraystretch}{1.50}  
\centering \fontsize{7}{5}\selectfont

\caption{Training Setup and Parameters}
\label{tab:parameters}
\begin{tabular}{|l|l|} 
 \hline                
Total Number of Video Clips          & 500              \\ \hline
\#Safe/Collision Scenario           & 385/115              \\ \hline
\ Max. Video Length (N), FPS       & 64 Frames, 20               \\ \hline
V-Jepa Encoder            & ViT-H            \\ \hline
Attentive Probe           & K400             \\ \hline
Original Frame Size $(W_o\times H_o)$   & 2048x2048          \\ \hline
Input Frame Size (WxH)    & 384x384          \\ \hline
Patch Size (pXp)            & 16x16            \\ \hline
Embedding Dimension (D)   & 1280             \\ \hline
Output of Classifier & 2                \\ \hline
\#Epochs, Optimizer                   & 40, AdamW              \\ \hline
Initial Learning Rate     & 0.001            \\ \hline
Modulation Schemes/SNR & BPSK/12dB, QAM-16/22dB \cite{11192791}         \\ \hline
Encoding Format& INT8, FP16, FP32                \\ \hline
Bandwidth & 20 MHz                \\ \hline
\end{tabular}
\end{table}

\subsection{Numerical Results}

\begin{figure}
    \centering
    \includegraphics[width=0.85\linewidth]{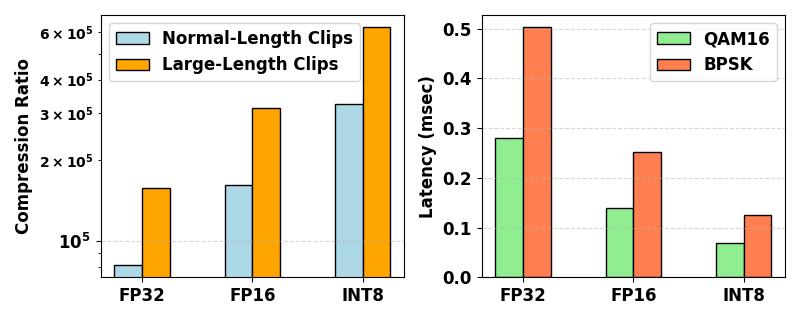}
    \caption{Average compression ratio for different video lengths and Tx latency for different modulation schemes \vspace{-0.2in}}
    \label{fig:comRes}
\end{figure}

From a communication perspective, two key metrics are analyzed: compression ratio and Tx latency (\figurename~\ref{fig:comRes}). 

The left portion of \figurename~\ref{fig:comRes} illustrates the achieved compression ratios, while the right portion depicts the corresponding Tx latencies. The raw videos obtained from different experiments vary in duration, resulting in different overall data sizes. To assess compression efficiency, the videos are categorized into two duration groups: normal-length and large-length clips. For each group, compression ratios are computed across the three encoding formats (FP32, FP16, and INT8), as described in Section~\ref{section:embedding}. In both cases, INT8 achieves the highest compression ratio, exceeding $10^{5}$, followed by FP16 and FP32. Large-length clips (64 frames) yield higher ratios than normal-length clips (33 frames), reflecting reduced transmission payload with longer videos. For example, assuming $N=64$, $H_o=W_o=2048$,  the raw video size is approximately $64\times2048\times2048\times3=0.81$~GB, while the transmitted semantic embeddings require only $1\times1280\times b=0.005$~MB, $0.0025$~MB, and $0.0012$~MB for FP32, FP16, and INT8, respectively, corresponding to compression gains of $1.6\times10^{5}$, $3.2\times10^{5}$, and $6.4\times10^{5}$.

To evaluate latency and Tx data rate, the SNR values for BPSK and QAM16 are set to 12~dB and 22~dB, respectively, as identified optimal previously~\cite{11192791}. QAM16 achieves the lowest Tx latency across all encodings, 0.27, 0.13, and 0.06~ms for FP32, FP16, and INT8,while BPSK yields 0.50, 0.25, and 0.12~ms under the same conditions. All values remain well below the 5~ms V2X threshold, confirming the framework’s suitability for real-time, low-latency vehicular communication.

\begin{figure}
    \centering
    \includegraphics[width=0.7\linewidth]{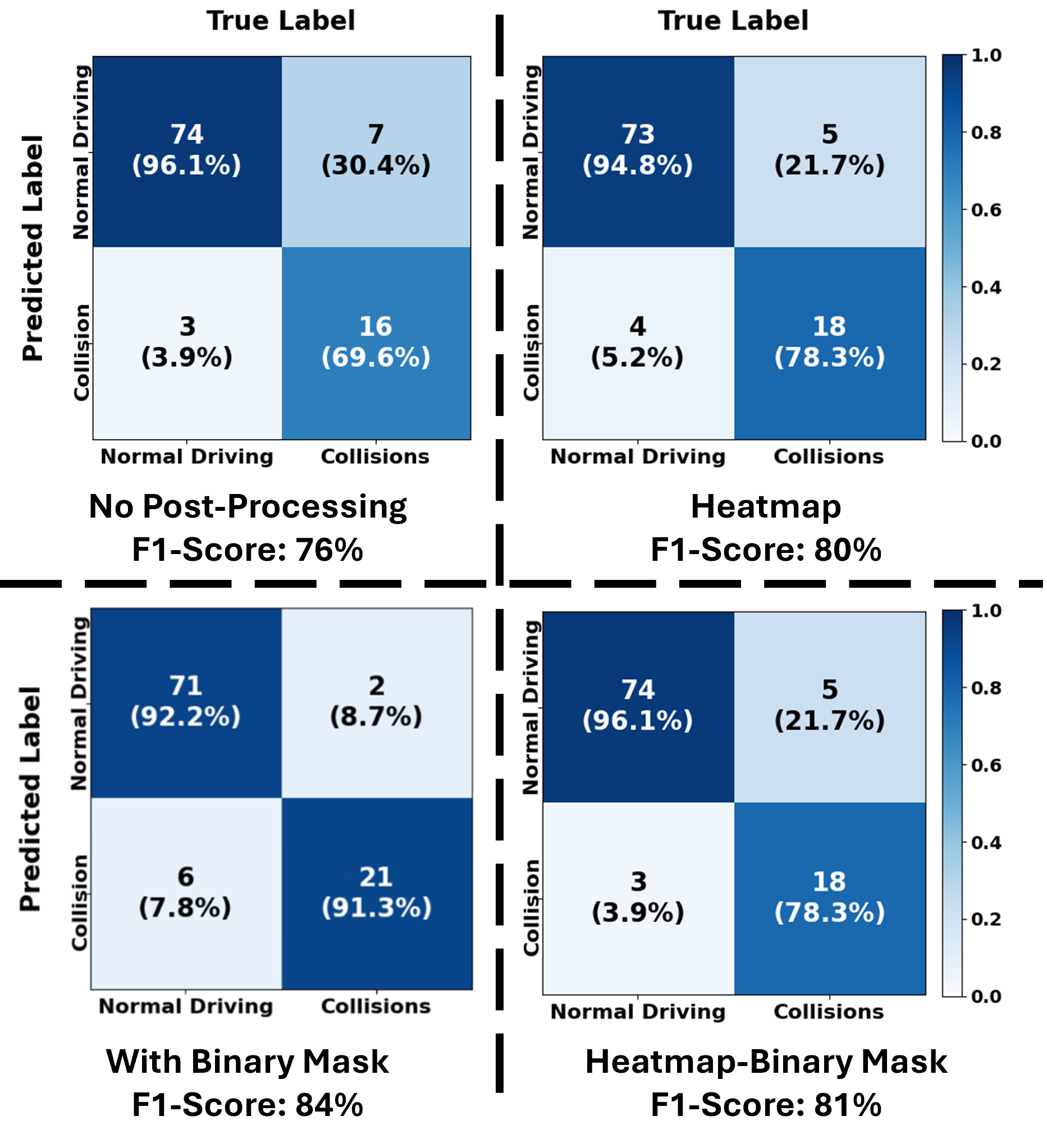}
    \caption{Results for Different Post-Processing Techniques}
    \label{fig:5}
\end{figure}

The evaluation of the V-JEPA model for traffic collision avoidance is conducted in two stages. First, various post-processing methods (Section~\ref{post-processing}) are analyzed to identify the configuration that provides the highest collision prediction accuracy. Next, the selected post-processing approach is used to adjust the frame gap preceding the collision to assess how early and reliably potential collisions can be detected. The resulting confusion matrices for different post-processing methods and the original clips are presented in \figurename~\ref{fig:5}.

\figurename \ref{fig:5} presents the confusion matrices for no post-processing (upper left), heatmap (upper right), binary masked (bottom left), and heatmap+binary masked (bottom right). For collision scenarios, each video clip was trimmed  to exclude 8 frames depicting or directly preceding the collision event. Model achieves 76\% F1 scores without post-processing, 80\% heatmap, 84\% binary masked, and 81.81\% heatmap+binary masked. Heatmap+binary masked  has a lower false positive than  binary masked, but binary masked has the highest F1 scores due to its low false negatives. This indicates that YOLO-based vehicle highlighting compromises performance in false negative cases, despite most collisions being detected properly. In the second step, we examine how the number of removed frames (frame-gap) before collisions affects collision prediction performance of V-JEPA Model (see Table \ref{table_result}).

\begin{table}[]  
\centering 
\caption{\centering Results for Different Previous-Frames Gap } \vspace{0.5em} 
\fontsize{7.5}{6.5}\selectfont 
\begin{tabular}{l|c|c|c}
                   & \textbf{\begin{tabular}[c]{@{}c@{}}12-Frames \\ Before Collision\end{tabular}} & \textbf{\begin{tabular}[c]{@{}c@{}}8-Frames \\ Before Collision\end{tabular}} & \textbf{\begin{tabular}[c]{@{}c@{}}4-Frames \\ Before Collision\end{tabular}} \\ \hline
\textbf{Accuracy}  & 90\%                                                                           & 92\%                                                                          & 90\%                                                                          \\ \hline
\textbf{Precision} & 60.9\%                                                                         & 91.3\%                                                                        & 82.6\%                                                                        \\ \hline
\textbf{Recall}    & 93.3\%                                                                         & 77.8\%                                                                        & 76.0\%                                                                        \\ \hline
\textbf{F1-score}  & 73.8\%                                                                         & 84.0\%                                                                        & 79.1\%                                                                        \\ \hline
\end{tabular}
\label{table_result}
\end{table}

The number of removed frames before the collisions indicates how early we can detect the collision. If the frame-gap before collision increases, the model tries to estimate the likelihood of future collision earlier. To achieve better results, we set the frame-gap to 12 (left), 8 (middle), and 4 (right) with binary mapped clips, resulting in 73.68\%, 84\%, and 79\% F1 scores, respectively. With 12 frames removed before the collision, the model yields highly false negative values because many collisions occur during cornering, and a longer frame cut results in clips ending before cornering begins. The 4-frame gap has a slightly lower F1 score than 8-frame gap before colliding because the model is already cornering and 4-frame removing used longer clips, making the model's work challenging.  


\section{Conclusions} \label{sec:6}
This paper has presented a V2X SemComm framework for proactive collision prediction using the V-JEPA. RSUs serve as semantic encoders, generating predictive spatiotemporal embeddings from video data captured via the QLabs digital twin. Post-processed clips highlighting critical regions were used with pretrained V-JEPA encoders and K400 attentive probes to convert frame sequences into compact vectors capturing essential motion cues, achieving 92\% accuracy and 8\% F1-score improvement in collision prediction. By transmitting low-dimensional embeddings instead of raw video, the framework reduces communication payloads by up to five orders of magnitude, enabling bandwidth-efficient and low-latency semantic message exchange. We are currently exploring cross-vehicle embedding fusion to enhance cooperative perception and scalability in next-generation V2X systems while working on the real dataset in more complex scenarios.

\section*{Acknowledgment }\label{Section6}
This work is supported in part by the MITACS Accelerate Program, in part by the NSERC CREATE TRAVERSAL program, and in part by the Ontario Research Fund-Research Excellence program under grant number ORF-RE012-026. The authors would like to thank Quanser for their support in the generation of the traffic data via Quanser Interactive Lab.


\bibliographystyle{IEEEtran}


\end{document}